\documentclass[10pt,twocolumn,letterpaper]{article}

 \usepackage{cvpr}              

\usepackage[dvipsnames]{xcolor}

\usepackage{graphicx}
\usepackage{graphicx}
\usepackage{amsmath}
\usepackage{amssymb}
\usepackage{booktabs}
\usepackage{amsmath}
\usepackage{amssymb}
\usepackage{bm}
\usepackage{booktabs}
\usepackage{multirow}
\usepackage{booktabs}
\usepackage{caption}
\usepackage{float}
\usepackage{algorithmic}
\usepackage{algorithm}
\usepackage{bbding}
\usepackage{array}
\usepackage{color}
\usepackage{diagbox}
\usepackage{threeparttable}
\usepackage{pifont}
\usepackage{tikz}
\usepackage{amsmath,amssymb} 
\usepackage{color}
\usepackage{amssymb}%
\usepackage{comment}
\usepackage[accsupp]{axessibility}  %
\definecolor{cvprblue}{rgb}{0.21,0.49,0.74}
\usepackage[pagebackref,breaklinks,colorlinks,citecolor=cvprblue]{hyperref}

\def\paperID{*****} 
\def\confName{CVPR}
\def\confYear{2024}

\title{Text-Animator: Controllable Visual Text Video Generation}

\author{Lin Liu\textsuperscript{1} \hspace{3mm}
    Quande Liu\textsuperscript{2} \hspace{3mm}
    Shengju Qian\textsuperscript{2} \hspace{3mm}
   Yuan Zhou\textsuperscript{3} \\ 
Wengang Zhou\textsuperscript{1} \hspace{3mm}
Houqiang Li\textsuperscript{1} \hspace{3mm}
Lingxi Xie\textsuperscript{4} \hspace{3mm}
Qi Tian\textsuperscript{4}\\
\\
 $^1$ EEIS Department,
University of Science and Technology of China \\
 $^2$ Tencent \quad
 $^3$Nanyang Technical University \\
 $^4$Huawei Tech.}

\begin{document}


\twocolumn[{%
\maketitle
\begin{figure}[H]
  \hsize=\textwidth 
   \centering
  \includegraphics[width=0.97\textwidth]{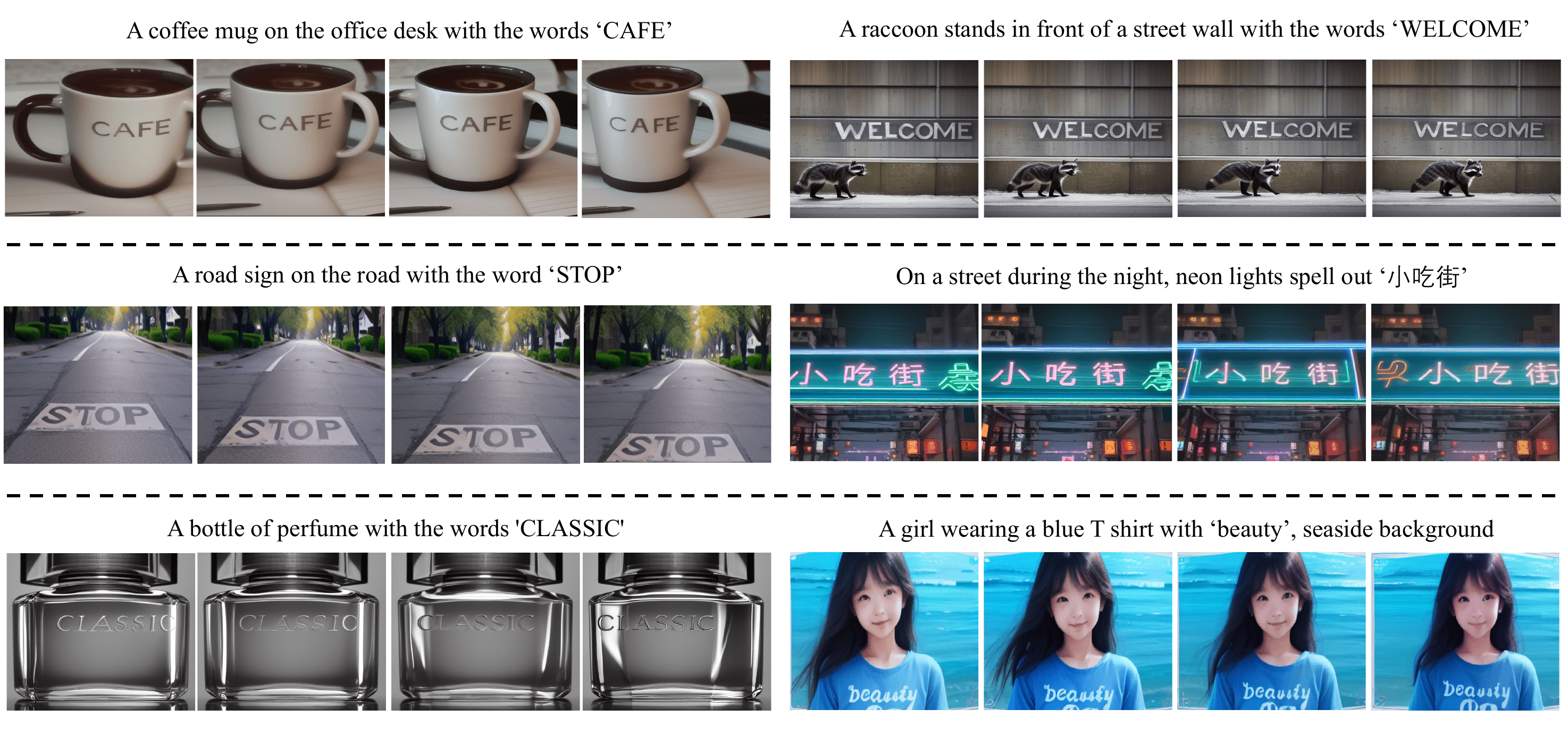}
  \caption{Given a sentence with visualized words, our Text-Animator is able to produce a wide range of videos that not only show the semantic information of given text prompts, but further align with the visualized words. Our method is a one stage method without further tuning.
}
  \label{fig:teaser}
\end{figure}
}]

\begin{abstract}

Video generation is a challenging yet pivotal task in various industries, such as gaming, e-commerce, and advertising. One significant unresolved aspect within T2V is the effective visualization of text within generated videos.
%
Despite the progress achieved in Text-to-Video~(T2V) generation, current methods still cannot effectively visualize texts in videos directly, as they mainly focus on summarizing semantic scene information, understanding, and depicting actions.
While recent advances in image-level visual text generation show promise, transitioning these techniques into the video domain faces problems, notably in preserving textual fidelity and motion coherence. 
In this paper, we propose an innovative approach termed Text-Animator for visual text video generation. 
Text-Animator contains a text embedding injection module to precisely depict the structures of visual text in generated videos. Besides, we develop a camera control module and a text refinement module to improve the stability of generated visual text by controlling the camera movement as well as the motion of visualized text.
Quantitative and qualitative experimental results demonstrate the superiority of our approach to the accuracy of generated visual text over state-of-the-art video generation methods.
The project page can be found in \url{laulampaul.github.io/text-animator.html}.

\end{abstract}    
\section{Introduction}
\label{sec:intro}

Video generation has become an important cornerstone in content-based generation, and has huge potential value in various domains including e-commerce, advertising, the film industry, etc.
%
For instance, in advertising scenarios, it is essential to display a manufacturer's specific logo or slogan in the generated video in the form of text, while also seamlessly integrating text with the products featured in the video (e.g. a piece of clothing). 
However, in current video generation approaches, the visualization of text/words in the generated video remains a challenging yet unresolved issue.
For example, in the first example of Fig.~\ref{fig:teaser}, we need to engrave the word "CAFE" on the mug and ensure that the movement of the text and the mug appear seamless and harmonious in the video.

%

Current T2V methods are unsuitable for these settings, as they typically focus on understanding the semantic-level information from a given prompt rather than interpreting specific words themselves.
For instance, given a text input as ``a person walking on the road," current T2V models can interpret the scene and produce a corresponding video about a person who walks on the road. However, these models fail to understand prompts at a more granular level. If the text input is modified to ``a person walking on the road, wearing a T-shirt with the word 'Hello World' printed on it," the generated results of current methods are far from satisfactory, due to their inability to accurately interpret the generation of the texts 'Hello World' and incorporate its associated motion information effectively.

%
Recently, some preliminary efforts have been made in the field of visual text generation, specifically in the paradigm of Text-to-Image (T2I) generation~\cite{yang2024glyphcontrol,tuo2023anytext,ma2023glyphdraw}. These trials have shown promising results, but they are only limited to the image domain. When extending this task to video scenarios, an intuitive approach is to use images generated by these methods as input for cutting-edge image-to-video (I2V) methods. However, most current I2V methods either focus on learning motion patterns in simply natural scenes~\cite{wang2024animatelcm,blattmann2023stable,dai2023animateanything} or deliberately omit data that include visual texts during dataset collection~\cite{blattmann2023stable}. As a result, videos generated by these methods fall into a dilemma generally called text collapse, which means that as the number of frames increases, the visualized text becomes increasingly blurry or loses its original structure (as demonstrated in Sec. \ref{sec:exp} of this paper). Therefore, it is difficult to directly extend visual text generation models from the image domain to the video domain.


Based on the above observations, we propose an effective solution for visual text video generation, which can effectively under texts for the description of videos and visual texts that should be generated. 
Our method not only reflects the semantics of the complete text, but also understands the fine-grain semantics of the input vocabulary, and effectively aggregates the two in terms of content while maintaining a good motion association (unable to visualize the movement of text and other content).

To achieve these goals, we propose a novel method called \textbf{Text-Animator}. Different from previous T2V methods, Text-Animator contains a text embedding injection module to enhance its precise understanding and generation capacity for visual text. Besides, a unified controlling strategy with camera and text position control is designed to improve the stability of the movement of visualized text and image content, thereby achieving unity and coordination of the text movements. Specifically, for camera control, the control information is applied to the main body of the network by considering the features of the camera's motion trajectories. The position control aims at controlling the specific position and size of visual text generated in videos. Owing to the comprehensive controlling strategy over the developed text embedding injection module, Text-Animator shows a superior capacity to generate stable and accurate visual text content in videos. 

In summary, the contributions of this paper can be concluded below:
\begin{itemize}
    \item We propose Text-Animator, a novel approach that can generate visual text in videos and maintain the structure consistency of generated visual texts. To our knowledge, this is the first attempt at the visual text video generation problem.
    
    \item We develop a text embedding injection module for Text-Animator that can accurately depict the structural information of visual text. Besides, we also propose a camera control and text refinement module to accurately control the camera movement and the motion of the generated visual text, to improve the generation stability. 

    \item Extensive experiments demonstrate that Text-Animator outperforms current text-to-video and image-to-video generation methods by a large margin on the accuracy of generated visual text.
\end{itemize}

\section{Related Work}
\label{sec:related}

\subsection{Visual Text Generation}
The goal of visual text generation is to integrate user-specified texts into images or videos and produce well-formed and readable visual text, therefore effectively ensuring that the texts fit well with the corresponding image content.
Current research mainly focuses on how to design an effective text encoder and considers the better guidance of text-conditioned controlling information.
For text encoder, as large language models develop~\cite{radford2021learning,raffel2020exploring}, it is a promising idea to directly use these models to encode text. However, this roadmap inevitably results in overlooking the character features of texts.
Recently, some works have optimized text encoders for character features.
%
GlyphDraw~\cite{ma2023glyphdraw} fine-tuned the text encoder for Chinese images for glyph embeddings. Chen et al.~\cite{chen2024diffute} trained a glyph-extracted image encoder for image editing. AnyText~\cite{tuo2023anytext} utilizes pretrained recognition model, PP-OCRv3~\cite{ppocrv3} for encoding text. 

To generate characters more accurately, control information of the text is required as additional input.
GlyphDraw~\cite{ma2023glyphdraw} uses explicit glyph images as conditions to render characters. GlyphControl~\cite{yang2024glyphcontrol} and AnyText~\cite{tuo2023anytext} embed text conditions in the latent space by the combination of the positions of text boxes and the rendered glyphs. different from \cite{ma2023glyphdraw,tuo2023anytext,yang2024glyphcontrol}, Yang et al.~\cite{yang2024glyphcontrol} use character-level segmentation masks as conditioned controlling information, allowing for finer granularity control.
To our knowledge, current methods mainly focus on addressing the visual text generation problem in the text-to-image domain, which cannot be utilized to tackle text-to-video visual text generation.
In this paper, we first explore the visual text generation task in the video domain.

\subsection{Video Generation}
Sora~\cite{videoworldsimulators2024}, a recent famous video generation model has attracted much attention from both the community of both industry and academia.
Before the emergence of diffusion-based models, lots of effort in this
field have been paid on methods based on GANs~\cite{goodfellow2014generative} or VQVAE~\cite{van2017neural,qian2023strait}.
Among these methods, the pre-trained Text-to-Image (T2I) model CogView2~\cite{ding2022cogview2} is utilized in CogVideo~\cite{ding2022cogview2} as the backbone, to enable generating long sequence videos in an auto-regressive way.
Based on autoregressive Transformers, NUWA~\cite{wu2022nuwa} combines three tasks, which are T2I, T2V, and video prediction.

Currently, diffusion models have become the mainstream method in video generation.
%
Make-A-Video~\cite{make} proposes to learn visual-textual correlations and thus capture video motion from unsupervised data. Some methods~\cite{an2023latent,wang2023modelscope,he2022latent,zhou2022magicvideo} design effective temporal modules to reduce computational complexity and model temporal relationships effectively. 
Multi-stage approaches~\cite{wang2023lavie,zhang2023show,saharia2022photorealistic} design models to be used in stages for achieving high-definition video generation.
These methods highlight the versatility and efficacy of diffusion models in advancing video generation capability.

\subsection{Controllable video generation}
In addition to conventional T2V models, some methods focus on making video generation controllable. 
In these methods, \cite{zhao2023motiondirector,wu2023lamp,wu2023tune,he2024id} turn to refer to specific video templates for controlling motion. 
However, despite the effectiveness of these methods in motion controlling, they typically require training new models on each template or template set, limiting their capability in controllable video generation.
Besides, VideoComposer~\cite{wang2024videocomposer} proposes to use motion vectors to control the video motion.
MotionCtrl~\cite{wang2023motionctrl} designs two control modules for camera motion and object motion control.
Drag-NUWA~\cite{yin2023dragnuwa} uses trajectories and text prompts in a joint way for video generation conditioned on an initial image. 
Different from these approaches, a dual control visual text generation model is utilized in our Text-Animator, where camera pose information
and position trajectories can effectively control the motion of videos and make the generation process more stable.


\section{Method}
In this section, we first introduce the pipeline of our Text-Animator in Sec. \ref{sec:pipeline}. Then, the details of the key components are introduced in Sec. \ref{sec:pcm}, Sec. \ref{sec:ccm}, and Sec. \ref{sec:vpr} respectively.

\begin{figure*}[h]
  \centering
  \includegraphics[width=0.97\textwidth]{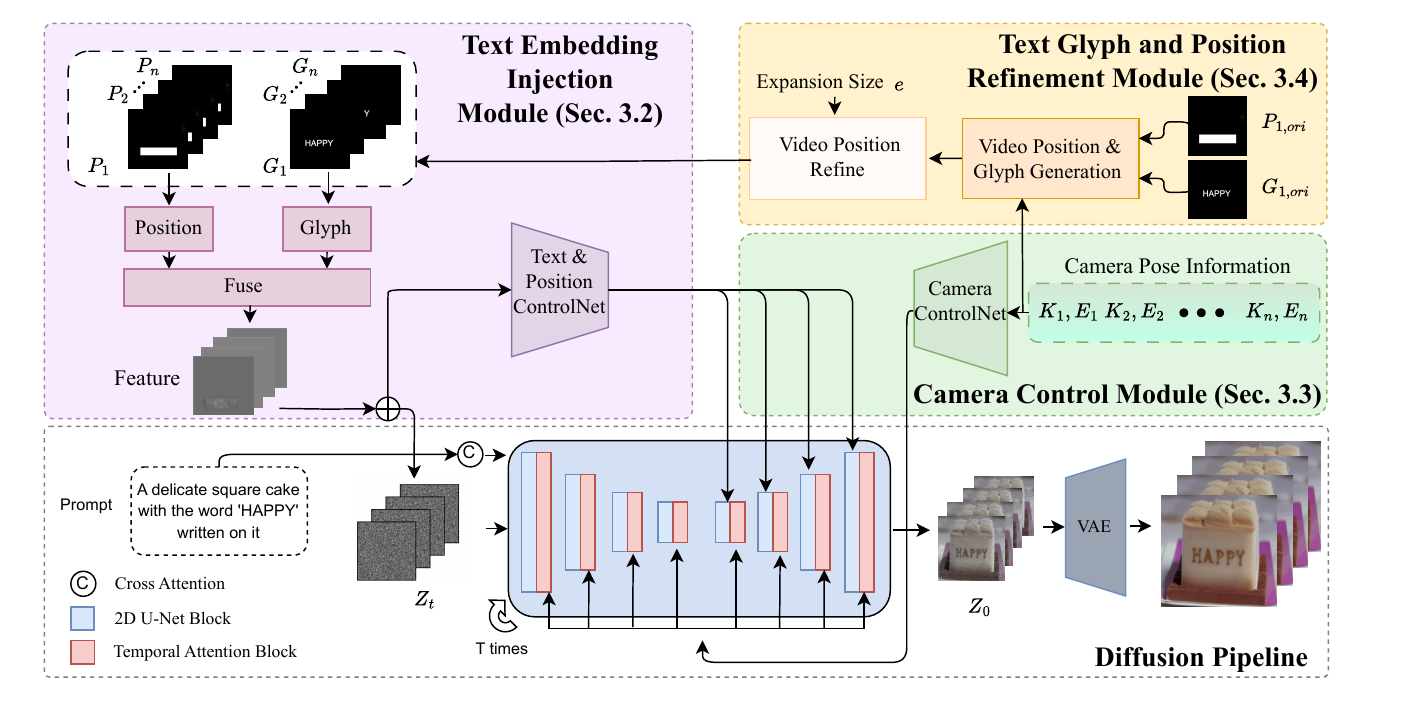}
  \caption{Framework of Text-Animator. Given a pre-trained 3D-UNet, the camera ControlNet takes camera embedding as input and outputs camera representations; the text and position ControlNet takes the combination feature $z_{c}$ as input and outputs position representations
These features are then integrated into the 2D Conv layers and temporal attention layers of 3D-UNet at their respective scales.}
  \label{fig:architecture}
\end{figure*}

\subsection{Text-conditioned 
Video Generation Pipeline}
\label{sec:pipeline}
Firstly, we introduce the overall framework of our network, as shown in Fig.~\ref{fig:architecture}. Our method consists of four parts that are  Text Embedding Injection Module, Camera Control Module, Text Glyph and Position Refinement Module, and 3D-UNet Module.
Given the integrated texts $T_{in}$, position map $P_{1,ori}$, and Camera Pose Information ($K_{1}, E_{1}, K_{2}, E_{2}, ... ,K_{n}, E_{n}$), the glyph map $G_{1,ori}$ is generated by rendering texts $T_{in}$ using a uniform font style onto an image based on their locations.
Then, the video position maps $P_{1},...,P_{n}$ and glyph maps $G_{1},...,G_{n}$ are generated by warping the $P_{1,ori}$ and $G_{1,ori}$ using camera pose information. 
Camera Control Module and Text Embedding Injection Module output multi-scale control features corresponding to their control input respectively.
The noise $z_t$ is fed into a 3D-UNet (the architecture of the 3D-UNet used in our work is as same as that used in AnimateDiffv3~\cite{guo2023animatediff}) to obtain an output $\epsilon_{t}$. In the inference stage, this output is passed through the decoder of the VAE to obtain the final output videos.

Recently, diffusion models have served as a primary framework for T2V generation and yielded promising results.
Current T2V methods are derived from the original formulations of diffusion models used in image generation.
%
More specifically, we generate the latent representation $z_{0}$ by applying Variational Autoencoder (VAE)~\cite{kingma2014auto} on the input video $x_{0}$.
Then, a sequence of $N$ latent features of $z_0$ is gradually perturbed with noise $\epsilon$ from a normal distribution over $T$ steps. Given the noisy input  $z_t$, a neural network $\hat{\epsilon}_{\theta}$ is trained to predict the added noise. 
In our work, we inject dual control signals (position control and camera control) into the denoising process, strengthening the stability of video generation. Specifically, these two control features are first fed into additional ControlNet $N_{p}$ and $N_{c}$ respectively, then injected into the generator through various operations. Hence, the objective of training our encoder is shown below:

\begin{equation}
\mathcal{L}=\mathbb{E}_{z_0, \epsilon, c_t,s_p, s_c t}\left[\| \epsilon-\hat{\epsilon}_\theta\left(z_t, c_t, N_p\left(s_p\right),N_c\left(s_c\right), t\right)\right] ,
\end{equation}
where $c_{t}$ is the embeddings of the corresponding text prompts, $s_p$ is the set of the position maps and glyph maps, and $s_c$ is the set of camera pose information ($s_{c}=K_{1}, E_{1}, K_{2}, E_{2}, ... ,K_{n}, E_{n}$).

\subsection{Text Embedding Injection Module}
\label{sec:pcm}
In the generation of videos with visual text, the first consideration is how to effectively embed the visual features of the required text into the base model (the pre-trained UNet model).
Inspired by previous methods of visualizing text in images~\cite{yang2024glyphcontrol,tuo2023anytext}, we embed text conditions in the latent space by combining the positions of text boxes and the rendered glyphs. Text boxes indicate the positions in the generated image where rendering should occur, while the rendered glyphs utilize existing font style (i.e., `Arial Unicode') to pre-initialize the style of the characters.
In addition, unlike image generation, video generation involves processing features across multiple frames. To leverage the pre-trained feature extractor used in image generation, we extract features from each frame using a frame-wise feature extractor, and then concatenate these features before feeding them into a pre-trained UNet model.

From the top left of Fig. \ref{fig:architecture}, we can see that the input to the position and glyph control module is the position map $P_{1}, P_{2}, ... , P_{n}$ and glyph map $G{1}, G_{2}, ... , G_{n}$ generated by the module in Sec. \ref{sec:vpr}.
We extract features of glyphs and positions separately using glyph convolution blocks and position convolution blocks, respectively.
Then, we merge these features using a fusion convolution block.
Finally, after combining these features with the noisy input $Z_t$, they are inputted into the text and position ControlNet.
The text and position ControlNet output multi-scale feature maps $F^{P}_{k} $. Following the ControlNet~\cite{zhang2023adding}, we fuse these features into the intermediate block and upsampling blocks of the UNet network, where they are directly added to the corresponding features.

\subsection{Camera Control for Stable Text Generation}
\label{sec:ccm}
After incorporating the text embedding injection module, our method is now capable of generating visual text videos with text that moves following the scene. However, this text movement can sometimes become disconnected from the movement of objects within the video. For instance, in the prompt `A sign that says ‘STOP’,' the text part "STOP" might move to the right while the sign moves to the left. To generate more stable videos, additional control modules need to be designed.
Therefore, we propose to use camera pose information to control the movement of text and ensure consistency with the scene content. In this section, we will primarily discuss how to embed camera pose information into the underlying model. In the next section, we will explore how to relate camera pose information to the position and glyph maps discussed in Section \ref{sec:pcm}.

To effectively embed the camera pose information ($K_{1}, E_{1}, K_{2}, E_{2}, ... ,K_{n}, E_{n}$) into the camera ControlNet, followed ~\cite{sitzmann2021light,he2024cameractrl}, we use the plucker embedding. And we briefly introduce it as follows.
A point $(u, v)$ in the image plane is represented as $\textbf{p}_{u,v} = (\bf{o} \times \textbf{d}_{u,v}, \textbf{d}_{u,v}) \in \mathbb{R}^6$, where $\bf{o} \in \mathbb{R}^3 $ denotes the camera center in the world coordinate space and $\textbf{d}_{u,v} \in \mathbb{R}^3 $ represents a directional vector in world coordinate space, calculated using the formula $\mathbf{d}_{u, v}=\mathbf{R K}^{-\mathbf{1}}[u, v, 1]^T+\mathbf{t}$.
$\mathbf{R}$ and $\mathbf{t}$ refer to the rotation matrix and the translation vector, respectively.
Thus, the embedding can be expressed as $\textbf{P} \in \mathbb{R}^{6 \times n \times H \times W}$, where
$H$ and $W$ are the height and width for the frames and $n$ represents the frame number.

The camera ControlNet consists of four blocks, each of them comprising a residual-based convolutional block and a transformer-based temporal attention block, allowing the network to learn temporal relationships within the camera pose information. The network outputs multi-scale features, \(F^c_{k} \in \mathbb{R} ^ {(b \times h_{k} \times w_{k}) \times n \times c_{k}}\).
After obtaining multi-scale camera features, it's necessary to integrate these features into the 3D-UNet architecture of the T2V model. The image latent features \(z_k\) and the camera pose features \(F_k\) are directly fused through pixel-wise addition.

\subsection{Auxiliary Text Glyph and Position Refinement}
\label{sec:vpr}
To enable the collaboration between the camera control module and the text embedding injection module, it is necessary to use the camera position information from videos as guidance to generate the position map and glyph map of subsequent frames by considering the guidance from the first frame. The generation method is as follows.

Given the first frame's map (position map or glyph map) of the first frame, $\bf{M}$, the intrinsic parameters $\bf{K}$, and the transformation matrix $\bf{T_{1}}$ of the first frame, and the transformation matrix $\bf{T_{n}}$ of the n-th frame.
We first calculate the transformation matrix $\bf{T_{1to2}= T_{1}^{-1}T_{2}}$ from the first frame to the second frame, and build the projection matrix $\bf{P=KT_{1to2}}$. 
Next, the pixel coordinate matrix of the first frame is converted to three-dimensional points $\bf{Point_{cam1_{3d}}}$ in the camera coordinate system. 
Here, due to lacking depth information, we assume that rendered texts on the same line is at the same depth.
Then, the relative transformation matrix $\bf{T_{1to2}}$ is used to transform to the second frame camera coordinate system and project it back onto the pixel plane using $\bf{P}$, followed by a normalization operation.
After normalization, the projected coordinates are constrained within the image boundary and filled into the second frame image.

After generating the position and glyph maps from the video, we observed in experiments that the relative position and size of the position feature map have a certain impact on the final generation results. If the position feature map is smaller, it affects the diversity of generated text, resulting in visual text that does not harmonize well with the content in the video. Conversely, if the position feature map is larger, it may lead to generated text containing incorrect or repeated characters. Therefore, we design a position refinement module.
First, we extract the centroid of the initial position map $P_{n,ori}$ and render the glyph map $G_{n,ori}$ at specific positions. Then, we extract the convex hull of the glyph map and expand it by adding an expansion factor $e$ to generate a new position map $P_{n}$.

\section{Experiments}
\label{sec:exp}
\subsection{Implementation Details}
We choose the AnimateDiffV3~\cite{guo2023animatediff} as the base text-to-video (T2V) model. The weights of the model’s motion module are initialized with AnimateDiffV3~\cite{guo2023animatediff}. The weights of other parts are initialized with DreamShaper~\cite{dreamshaper} or original SD1.5~\cite{Rombach_2022_CVPR}.
Camera controlnet and text and position controlnet are trained using methods and datasets in \cite{he2024cameractrl} and \cite{tuo2023anytext}. Finally, all the parts are aggregated and the parameters are fixed for inference.
Image dimensions of $G$ and $P$ are set to be $1024 \times 1024$ and $512 \times 512$, respectively.
The expansion size $e$ is set to 1.2.
During the sampling process, we randomly selected some hint prompts (like `these texts are written on it: xxx') and concatenated them to the caption. 
The inference step and the guidance scale are set to 25 and 7.5, respectively.
Finally, the model outputs the videos with the size $16 \times 256 \times 384$.

\subsection{Dataset and Metrics}
Because of lacking the Text-to-video dataset for visual text generation evaluation, we use the LAION subsets of AnyText-benchmark~\cite{tuo2023anytext} for evaluating the effectiveness of visual text video generation. However, 
in this dataset, some images have text and main content separated, while others consist only of text without any image content, which is meaningless for video generation. Therefore, we selected about 90 images from the dataset to form the test set, which is named the LAION subset.

Firstly, we need to assess the accuracy and quality of text generation. According to the paper~\cite{tuo2023anytext}, we employed the Sentence Accuracy (Sen. Acc) metric, where each generated text line is cropped according to the specified position and fed into an OCR model to obtain predicted results. Additionally, the Normalized Edit Distance (NED)~\cite{marzal1993computation} is used to measure the similarity between two strings.
To ensure that our method has better video generation capabilities, we utilize the Fréchet Inception Distance (FID) to assess the video appearance quality between generated videos and real-world videos.
Moreover, we also adopted the Prompt similarity and the Frame similarity metric. The former evaluates the semantic similarity between the input description and output video, while the latter evaluates the continuity of the generated videos.

\subsection{Quantitative Results}
\begin{table*}[t]
  \centering
   \caption{Quantitative comparison results on the LAION-subset dataset. The best results are shown in Bold and the second best are underlined.}
  \resizebox{16.8cm}{!} {
  \begin{tabular}{c|cccccc}
    \toprule
      Method & Parameters & Sen. ACC $\uparrow$ & NED $\uparrow$& FID $\downarrow$ & Prompt similarity $\uparrow$& Frame similarity$\uparrow$\\
      \midrule
      Anytext~\cite{tuo2023anytext} + AnimateLCM~\cite{wang2024animatelcm} &  2726M &0.220 &\underline{0.615}&\bf{153.7} &33.62&75.91\\
      Anytext~\cite{tuo2023anytext}  + I2VGen-XL~\cite{zhang2023i2vgen}  & 2785M  &\underline{0.267}&0.582&184.9 & 30.18 & 79.74\\
      GlyphControl~\cite{yang2024glyphcontrol}  + AnimateLCM~\cite{wang2024animatelcm} & 2625M  & 0.139&0.303 & 182.3 & \bf{34.00} &76.03 \\
       GlyphControl~\cite{yang2024glyphcontrol} + I2VGen-XL~\cite{zhang2023i2vgen}& 2684M & 0.197 &0.298&186.1&32.26 &79.98\\
     \midrule
      Animatediff-SDXL (Text Lora A)~\cite{guo2023animatediff}  & 2927M  & 0.209  & 0.555 & 262.1 & 32.72 &74.22\\
       Animatediff-SDXL (Text Lora B)~\cite{guo2023animatediff}  & 2927M  & 0.197   & 0.528& 275.2  &32.51 & 78.10\\
       \midrule
       
      Ours &  1855M& \bf{0.779} & \bf{0.802}   & \underline{180.6} & \underline{33.78} & \bf{92.66}\\
    \bottomrule
  \end{tabular}
  }
  \label{tab:res1}
\end{table*}

\begin{table}[t]
  \centering
   \caption{Quantitative comparison results on the LAION-subset dataset with some T2V methods.}
  \resizebox{6.4cm}{!} {
  \begin{tabular}{c|cc}
    \toprule
      Method & Sen. ACC $\uparrow$ & NED $\uparrow$\\
      \midrule
       Open-SORA~\cite{opensora} & -- & 0.081\\
       Morph Studio~\cite{morph} & -- & 0.255 \\
       Pika~\cite{pikalabs} &0.267 & 0.611\\
       Gen-2~\cite{Gen-2} &0.279 &0.708\\
      
      Ours &   \bf{0.779} & \bf{0.802}  \\
    \bottomrule
  \end{tabular}
  }
  \label{tab:resothers}
\end{table}
The quantitative results are shown in Table \ref{tab:res1}. The compared methods are divided into two parts. The first part is the combination of the specific image visual text generation works (GlyphControl~\cite{yang2024glyphcontrol} and Anytext~\cite{tuo2023anytext}) + state-of-the-art I2V works (AnimateLCM~\cite{wang2024animatelcm}, I2VGEN-XL~\cite{zhang2023i2vgen}). 
The second part is the one-stage method. We use the Animatediff-SDXL as the base model and two finetuned lora weight from CIVIAI, denoted as Animatediff-SDXL (Text Lora A)\footnote{This lora model is from \url{ https://civitai.com/models/419492?\\modelVersionId=467355}} and Animatediff-SDXL (Text Lora B)\footnote{This lora model is from \url{https://civitai.com/models/221240/texta-generate-text-with-sdxl}} in Table~\ref{tab:res1} respectively. These two lora weight are finetuned using some images with visual text. 
From Table \ref{tab:res1}, we can see that the parameters of these methods are much larger than that of our method (over 41\%).
Moreover, our method significantly outperforms other methods in terms of the accuracy of generating visual text, as measured by evaluation metrics Sen. ACC and NED (leading by 191.8\% and 30.4\% respectively compared to the best method). This reflects the accuracy of the text generated by our method, and the text does not collapse in the generated videos.
As for the metric measuring the similarity between the generated video and the input text (FID and Prompt similarity), our method achieved the second-best result. In terms of Prompt Similarity, the gap with the best method is only 0.6\%.
In the metric measuring video stability and frame Similarity, our method achieved the second-best result. We observed that the best method, Pika, tends to generate videos with smaller movements, giving them an advantage in this metric.

Besides, in Table~\ref{tab:resothers}, we also compare with Open-SORA~\cite{opensora} and three recent SOTA API, Morph Studio~\cite{morph}, Pika~\cite{pikalabs}, and Gen-2~\cite{Gen-2}.
Open-SORA and Morph Studio do not have the Sen. ACC score because they cannot generate correct sentences or words.
Our method significantly outperforms other methods in terms of Sen. ACC and also performs better than other methods in NED.

\subsection{Qualitative Results}
\begin{figure*}[t]
  \centering
  \includegraphics[width=0.98\textwidth]{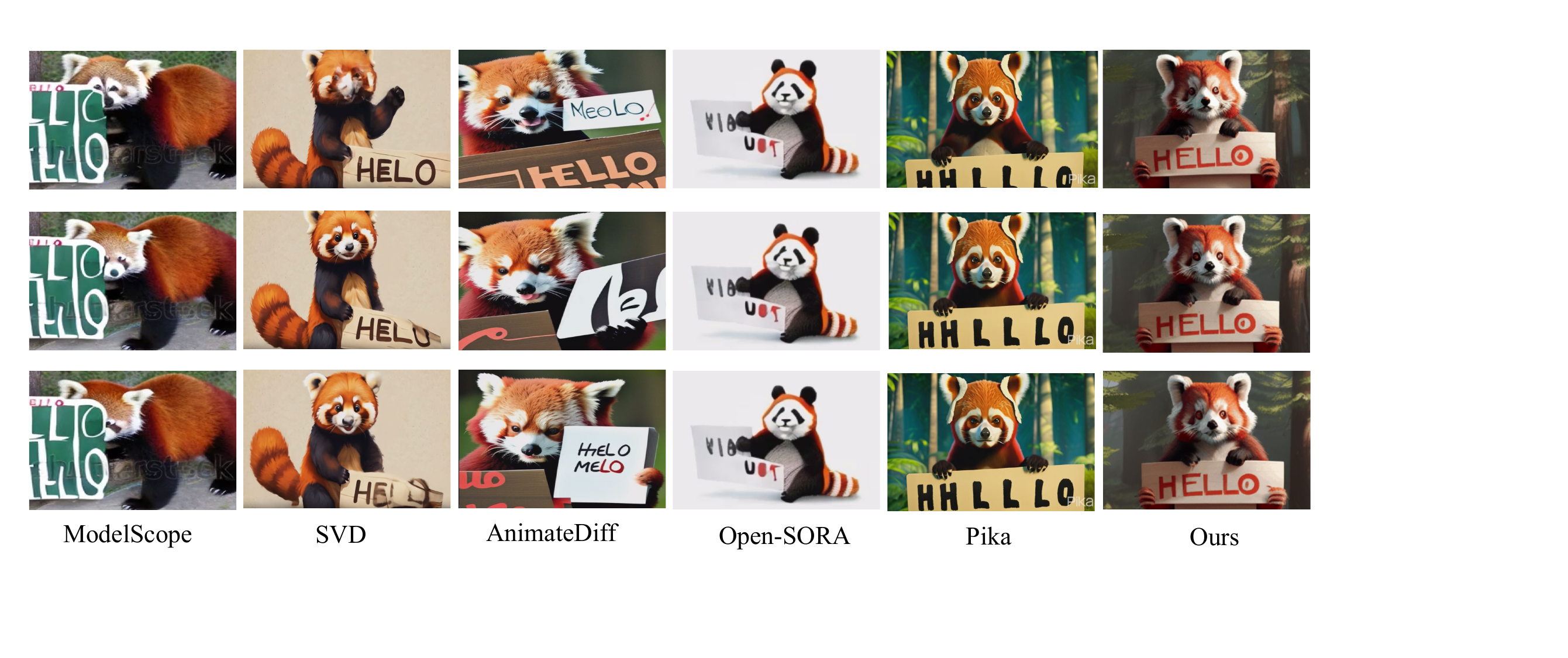}
  \caption{Qualitative comparison of Text-Animator and state-of-the-art T2V models or APIs in visual text generation. The prompt is `A red panda is holding a sign that says `HELLO''.}
  \label{fig:vis1}
\end{figure*}

In this subsection, we first compared our model with state-of-the-art T2V models or
APIs in the field of text-to-video generation (including ModelScope~\cite{wang2023modelscope} SVD~\cite{blattmann2023stable} (Stable Video Diffusion), AnimatedDiff~\cite{guo2023animatediff}, Open-SORA~\cite{opensora}, and Pika~\cite{pikalabs}) as shown in Fig. \ref{fig:vis1}. 
These models show the ability on context understanding, but they fail to generate specific texts and maintain textual consistency over time.
Compared to SVD, our model not only accurately renders each character (ours: `HELLO' vs SVD: `HELO' or Pika: `HHLLLO'), but also maintains consistency over time. SVD fails to learn the motion information of the text, causing the text to become increasingly disordered as time passes.

\begin{figure*}[t]
  \centering
  \includegraphics[width=0.96\textwidth]{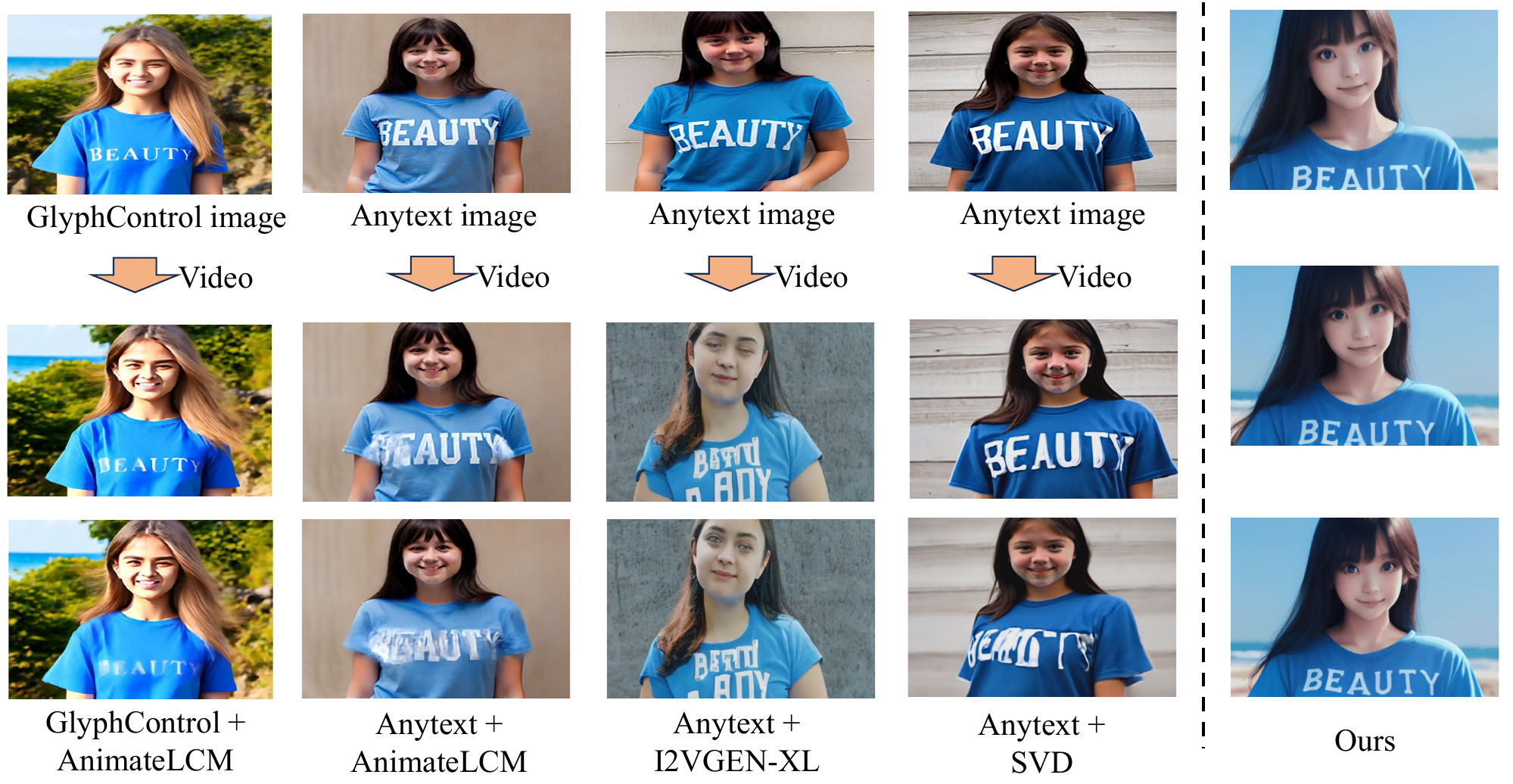}
  \caption{Qualitative comparison of Text-Animator and the combination of state-of-the-art T2I visual text generation models (GpyphControl and Anytext) and I2V models (AnimateLCM~\cite{wang2024animatelcm}, I2VGen-XL~\cite{zhang2023i2vgen}, and SVD). The prompt is `A girl wearing a blue T-shirt with the words `BEAUTY', slight smile, seaside background'.}
  \label{fig:vis2}
\end{figure*}

\begin{figure*}[t]
  \centering
  \includegraphics[width=0.96\textwidth]{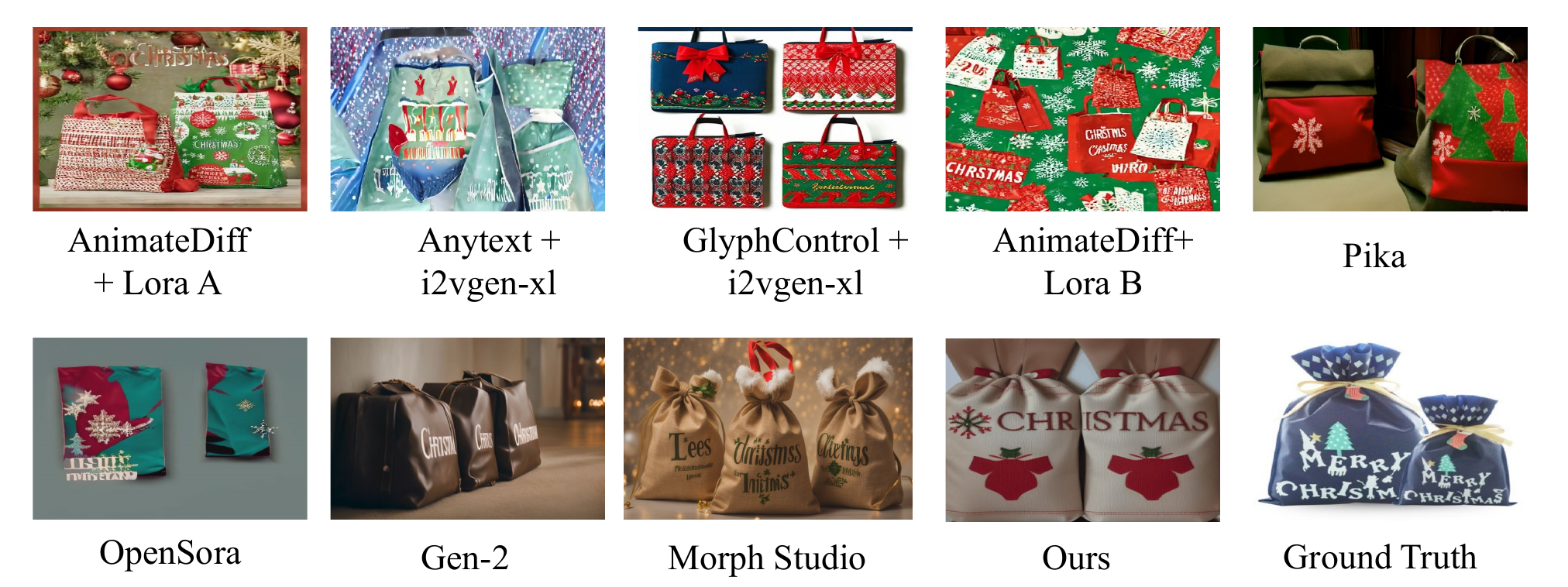}
  \caption{Qualitative comparison of Text-Animator and others on one example of the LAION-subset dataset. The prompt is `Two bags with the word 'CHRISTMAS' designed on it'. Other methods cannot generate the correct word (Please zoom to see the results).}
  \label{fig:vis5}
\end{figure*}

As for comparison with specific visual text generation works, since there is currently no T2V work specifically designed for visual text generation, we contrast our approach with methods combining specific T2I works for visual text generation (such as GlyphControl~\cite{yang2024glyphcontrol} and Anytext~\cite{tuo2023anytext}) and state-of-the-art I2V works (such as AnimateLCM~\cite{wang2024animatelcm}, I2VGen-XL, and SVD~\cite{blattmann2023stable}).
As shown in Fig.~\ref{fig:vis2}, our method shows superior integration of generated text with background, while Anytext cannot generate the seaside background.
When using I2V methods to generate videos from reference frame images, the text parts are often blurred or distorted. Our approach maintains the clarity of the text parts well and moves in coordination with the image content.
Besides, in Fig.~\ref{fig:vis5}, we show one example of the LAION-subset dataset. Only our method can correctly display the visual characters (CHRISTMAS) and the number of bags (two).

\begin{figure}[t]
  \centering
  \includegraphics[width=0.47\textwidth]{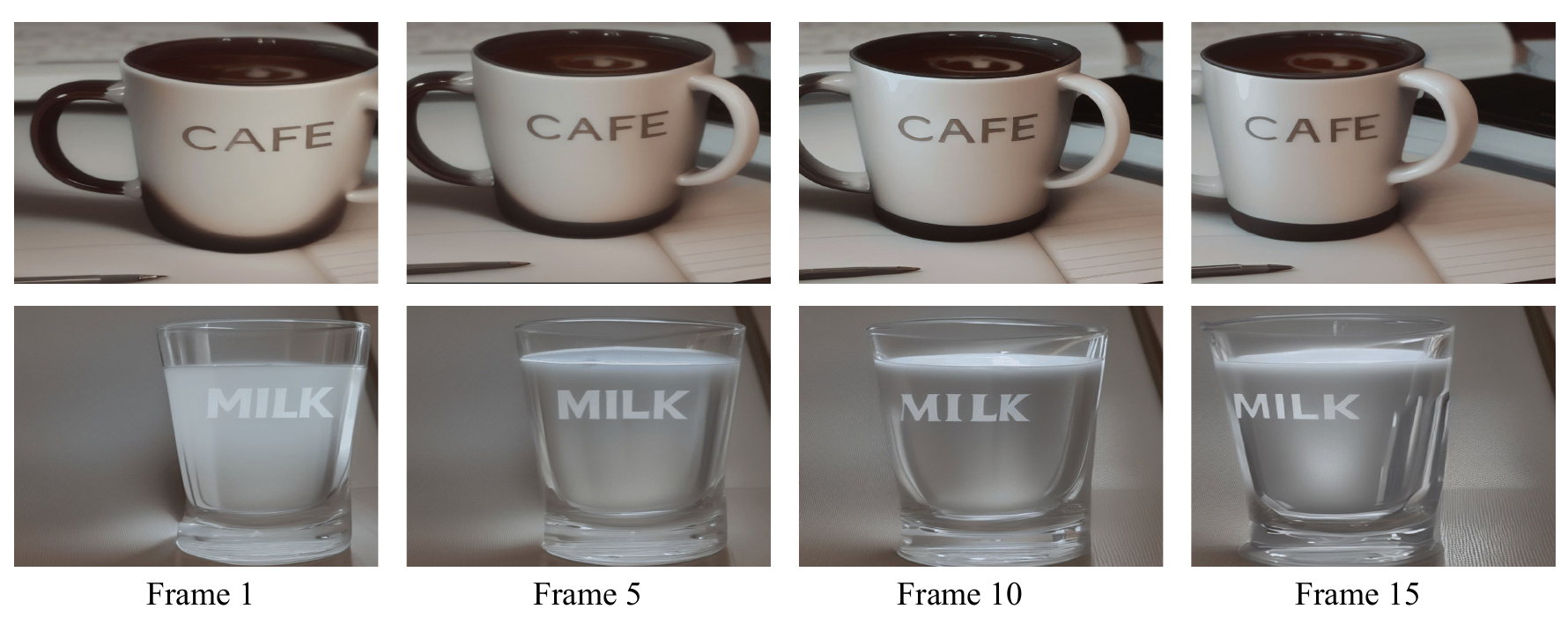}
  \caption{The example of large-area text movement, demonstrates that our method does not cause damage to the text when moving text over a large area.}
  \label{fig:vis3}
\end{figure}

\begin{figure}[t]
  \centering
  \includegraphics[width=0.47\textwidth]{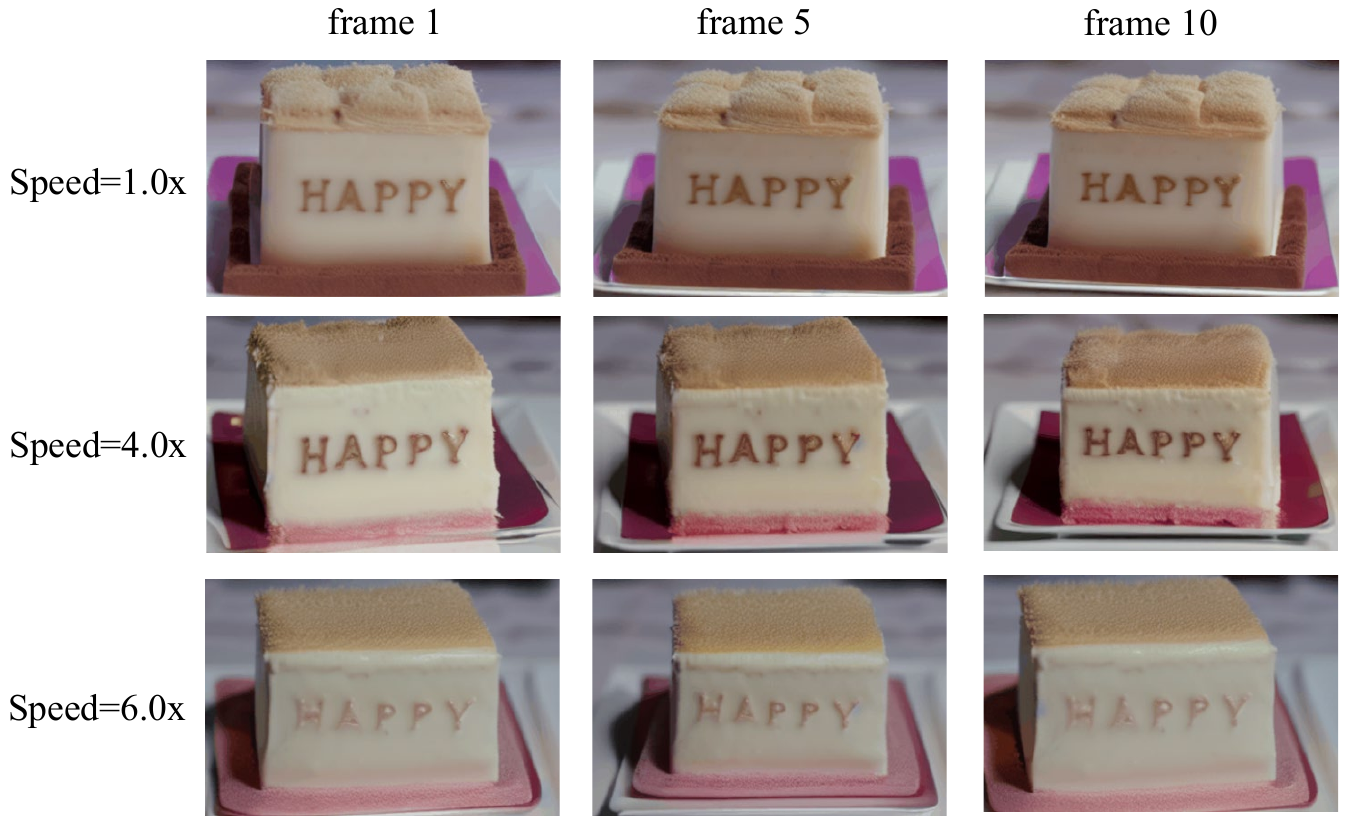}
  \caption{The comparison of the same text and camera information at different speeds. The prompt is `A delicious and square cake with the words `HAPPY''.}
  \label{fig:vis4}
\end{figure}

At the same time, we also conducted experiments to verify the robustness of our method. In Fig.~\ref{fig:vis3}, we demonstrate the robustness of our method for large movement in the text region. The existing SOTA methods deformed the text area during small movements (as shown in the example above), so the visualization results of these methods are not shown here. 
The texts for these two examples are `A coffee mug with the words `cafe' on the office desk' and `A bottom of milk with the words `MILK'. The direction of movement is from right to left. We can see that the structure of our text can still be maintained even with a large range of camera movements. In Fig.~\ref{fig:vis4}, we demonstrate that under the same camera information, we can control its movement speed by sampling the camera information of the interval frames. At a speed of 4 or 6 times the original speed, our method is still able to maintain the structure of the text.

\begin{table}[t]
  \centering
   \caption{Ablation studies on the LAION-subset dataset.}
  \resizebox{8.5cm}{!} {
  \begin{tabular}{c|ccc}
    \toprule
      Method  & Sen. ACC & NED$\uparrow$ & FID $\downarrow$ \\
      \midrule
      W/o camera control &0.755 &0.786& 183.2\\
      W/o position control &0.732 & 0.775 &185.7\\
       W/o position refinement&0.755 & 0.763 & 180.9\\
       Expansion size=0.9 & \bf{0.779} & \bf{0.804}  & 181.9\\
       Expansion size=1.4& 0.767 &  0.791 & 181.3\\
       \midrule
      Full model & \bf{0.779} & \underline{0.802}   & \bf{180.6} \\
    \bottomrule
  \end{tabular}
  }
  \label{tab:ab}
\end{table}

\subsection{Ablation Study}
In this part, to illustrate the contributions of our method, we conduct ablation studies on LAION-subset. The quantitative comparisons are shown in Table \ref{tab:ab}.

\textbf{Dual control}: We conduct an ablation
study to analyze the effectiveness of the dual control design.
Generally speaking, it is feasible to use only position boxes for guidance without using camera poses. Therefore, we designed the `W/o camera control' model, which removed the camera guidance module compared to the original model. In addition, we removed the position block and only used camera pose and glyph embedding, and named this model `W/o position control'.
In Table \ref{tab:ab}, we can see that on the xxx metric, the performance of the 'W/o camera control' model has decreased by 0.016 on NED compared to the original model, and the performance of the 'W/o position control' model has decreased by 0.027 on NED compared to the original model. 

\textbf{Position refinement and expansion size}:
 We also conduct experiments to analyze the effectiveness of our proposed refinement module. When the video position refinement is removed, we use the default position in the LAION subset. And we denote the model as `w/o Position Refinement' in Table \ref{tab:ab}. We can see that the original position will decrease the accuracy. Besides, we conduct experiments about the proper expansion size. 
 We tried two expansion coefficients: 0.9 (smaller than 1.2) and 1.4 (larger than 1.2). It can be observed that although the smaller expansion coefficient improves the accuracy of the text in the video, it negatively impacts the quality of the video generation. On the other hand, the larger expansion coefficient causes some characters to appear repeatedly in the video, thereby reducing the accuracy of the text.
 







\section{Conclusion}
In conclusion, this paper presents Text-Animator, an innovative approach to address the challenge of integrating textual elements effectively into generated videos within the visual text video generation domain. 
Text-Animator emphasizes not only semantic understanding of text but also fine-grained textual semantics, ensuring that visualized text is dynamically integrated into video content while maintaining motion coherence. Our approach introduces dual control mechanisms—camera and position control- to synchronize text animation with video motion, thereby enhancing unity and coordination between textual elements and video scenes.
Through extensive quantitative and visual experiments, we have demonstrated that Text-Animator outperforms existing T2V and hybrid T2I/I2V methods in terms of video quality and fidelity of textual representation.
Our contributions not only address current challenges but also inspire further exploration and innovation in this rapidly evolving field of multimedia content generation.
{
    \small
    \bibliographystyle{ieeenat_fullname}
    \bibliography{main}

\begin{thebibliography}{44}
\providecommand{\natexlab}[1]{#1}
\providecommand{\url}[1]{\texttt{#1}}
\expandafter\ifx\csname urlstyle\endcsname\relax
  \providecommand{\doi}[1]{doi: #1}\else
  \providecommand{\doi}{doi: \begingroup \urlstyle{rm}\Url}\fi

\bibitem[An et~al.(2023)An, Zhang, Yang, Gupta, Huang, Luo, and Yin]{an2023latent}
Jie An, Songyang Zhang, Harry Yang, Sonal Gupta, Jia-Bin Huang, Jiebo Luo, and Xi Yin.
\newblock Latent-shift: Latent diffusion with temporal shift for efficient text-to-video generation.
\newblock \emph{arXiv preprint arXiv:2304.08477}, 2023.

\bibitem[Blattmann et~al.(2023)Blattmann, Dockhorn, Kulal, Mendelevitch, Kilian, Lorenz, Levi, English, Voleti, Letts, et~al.]{blattmann2023stable}
Andreas Blattmann, Tim Dockhorn, Sumith Kulal, Daniel Mendelevitch, Maciej Kilian, Dominik Lorenz, Yam Levi, Zion English, Vikram Voleti, Adam Letts, et~al.
\newblock Stable video diffusion: Scaling latent video diffusion models to large datasets.
\newblock \emph{arXiv preprint arXiv:2311.15127}, 2023.

\bibitem[Brooks et~al.(2024)Brooks, Peebles, Holmes, DePue, Guo, Jing, Schnurr, Taylor, Luhman, Luhman, Ng, Wang, and Ramesh]{videoworldsimulators2024}
Tim Brooks, Bill Peebles, Connor Holmes, Will DePue, Yufei Guo, Li Jing, David Schnurr, Joe Taylor, Troy Luhman, Eric Luhman, Clarence Ng, Ricky Wang, and Aditya Ramesh.
\newblock Video generation models as world simulators.
\newblock 2024.

\bibitem[Chen et~al.(2024)Chen, Xu, Gu, Li, Meng, Zhu, Wang, et~al.]{chen2024diffute}
Haoxing Chen, Zhuoer Xu, Zhangxuan Gu, Yaohui Li, Changhua Meng, Huijia Zhu, Weiqiang Wang, et~al.
\newblock Diffute: Universal text editing diffusion model.
\newblock \emph{Advances in Neural Information Processing Systems}, 36, 2024.

\bibitem[Chenxia~Li(2022)]{ppocrv3}
Ruoyu Guo Xiaoting Yin Kaitao Jiang Yongkun Du Yuning Du Lingfeng Zhu Baohua Lai Xiaoguang Hu Dianhai Yu Yanjun~Ma Chenxia~Li, Weiwei~Liu.
\newblock Pp-ocrv3: More attempts for the improvement of ultra lightweight ocr system.
\newblock \emph{arXiv preprint arXiv:2206.03001}, 2022.

\bibitem[Dai et~al.(2023)Dai, Zhang, Yao, Qiu, Zhu, Qin, and Wang]{dai2023animateanything}
Zuozhuo Dai, Zhenghao Zhang, Yao Yao, Bingxue Qiu, Siyu Zhu, Long Qin, and Weizhi Wang.
\newblock Animateanything: Fine-grained open domain image animation with motion guidance.
\newblock \emph{arXiv e-prints}, pages arXiv--2311, 2023.

\bibitem[Ding et~al.(2022)Ding, Zheng, Hong, and Tang]{ding2022cogview2}
Ming Ding, Wendi Zheng, Wenyi Hong, and Jie Tang.
\newblock Cogview2: Faster and better text-to-image generation via hierarchical transformers.
\newblock \emph{Advances in Neural Information Processing Systems}, 35:\penalty0 16890--16902, 2022.

\bibitem[Gen-2(September 25, 2023)]{Gen-2}
Gen-2, September 25, 2023.
\newblock \url{https://research.runwayml.com/gen2}.

\bibitem[Goodfellow et~al.(2014)Goodfellow, Pouget-Abadie, Mirza, Xu, Warde-Farley, Ozair, Courville, and Bengio]{goodfellow2014generative}
Ian Goodfellow, Jean Pouget-Abadie, Mehdi Mirza, Bing Xu, David Warde-Farley, Sherjil Ozair, Aaron Courville, and Yoshua Bengio.
\newblock Generative adversarial nets.
\newblock \emph{Advances in neural information processing systems}, 27, 2014.

\bibitem[Guo et~al.(2023)Guo, Yang, Rao, Wang, Qiao, Lin, and Dai]{guo2023animatediff}
Yuwei Guo, Ceyuan Yang, Anyi Rao, Yaohui Wang, Yu Qiao, Dahua Lin, and Bo Dai.
\newblock Animatediff: Animate your personalized text-to-image diffusion models without specific tuning.
\newblock \emph{arXiv preprint arXiv:2307.04725}, 2023.

\bibitem[He et~al.(2024{\natexlab{a}})He, Xu, Guo, Wetzstein, Dai, Li, and Yang]{he2024cameractrl}
Hao He, Yinghao Xu, Yuwei Guo, Gordon Wetzstein, Bo Dai, Hongsheng Li, and Ceyuan Yang.
\newblock Cameractrl: Enabling camera control for text-to-video generation.
\newblock \emph{arXiv preprint arXiv:2404.02101}, 2024{\natexlab{a}}.

\bibitem[He et~al.(2024{\natexlab{b}})He, Liu, Qian, Wang, Hu, Cao, Yan, Zhou, and Zhang]{he2024id}
Xuanhua He, Quande Liu, Shengju Qian, Xin Wang, Tao Hu, Ke Cao, Keyu Yan, Man Zhou, and Jie Zhang.
\newblock Id-animator: Zero-shot identity-preserving human video generation.
\newblock \emph{arXiv preprint arXiv:2404.15275}, 2024{\natexlab{b}}.

\bibitem[He et~al.(2022)He, Yang, Zhang, Shan, and Chen]{he2022latent}
Yingqing He, Tianyu Yang, Yong Zhang, Ying Shan, and Qifeng Chen.
\newblock Latent video diffusion models for high-fidelity long video generation.
\newblock \emph{arXiv preprint arXiv:2211.13221}, 2022.

\bibitem[Kingma and Welling(2014)]{kingma2014auto}
Diederik~P Kingma and Max Welling.
\newblock Auto-encoding variational bayes.
\newblock \emph{ICLR}, 2014.

\bibitem[labs(2023)]{pikalabs}
Pika labs, 2023.
\newblock \url{https://www.pika.art/}.

\bibitem[Lykon(2023)]{dreamshaper}
Lykon, 2023.
\newblock \url{https://huggingface.co/Lykon/dreamshaper-8}.

\bibitem[Ma et~al.(2023)Ma, Zhao, Chen, Wang, Niu, Lu, and Lin]{ma2023glyphdraw}
Jian Ma, Mingjun Zhao, Chen Chen, Ruichen Wang, Di Niu, Haonan Lu, and Xiaodong Lin.
\newblock Glyphdraw: Learning to draw chinese characters in image synthesis models coherently.
\newblock \emph{arXiv preprint arXiv:2303.17870}, 2023.

\bibitem[Marzal and Vidal(1993)]{marzal1993computation}
Andres Marzal and Enrique Vidal.
\newblock Computation of normalized edit distance and applications.
\newblock \emph{IEEE transactions on pattern analysis and machine intelligence}, 15\penalty0 (9):\penalty0 926--932, 1993.

\bibitem[Qian et~al.(2023)Qian, Chang, Li, Zhang, Jia, and Zhang]{qian2023strait}
Shengju Qian, Huiwen Chang, Yuanzhen Li, Zizhao Zhang, Jiaya Jia, and Han Zhang.
\newblock Strait: Non-autoregressive generation with stratified image transformer.
\newblock \emph{arXiv preprint arXiv:2303.00750}, 2023.

\bibitem[Radford et~al.(2021)Radford, Kim, Hallacy, Ramesh, Goh, Agarwal, Sastry, Askell, Mishkin, Clark, et~al.]{radford2021learning}
Alec Radford, Jong~Wook Kim, Chris Hallacy, Aditya Ramesh, Gabriel Goh, Sandhini Agarwal, Girish Sastry, Amanda Askell, Pamela Mishkin, Jack Clark, et~al.
\newblock Learning transferable visual models from natural language supervision.
\newblock In \emph{International conference on machine learning}, pages 8748--8763. PMLR, 2021.

\bibitem[Raffel et~al.(2020)Raffel, Shazeer, Roberts, Lee, Narang, Matena, Zhou, Li, and Liu]{raffel2020exploring}
Colin Raffel, Noam Shazeer, Adam Roberts, Katherine Lee, Sharan Narang, Michael Matena, Yanqi Zhou, Wei Li, and Peter~J Liu.
\newblock Exploring the limits of transfer learning with a unified text-to-text transformer.
\newblock \emph{Journal of machine learning research}, 21\penalty0 (140):\penalty0 1--67, 2020.

\bibitem[Rombach et~al.(2022)Rombach, Blattmann, Lorenz, Esser, and Ommer]{Rombach_2022_CVPR}
Robin Rombach, Andreas Blattmann, Dominik Lorenz, Patrick Esser, and Bj\"orn Ommer.
\newblock High-resolution image synthesis with latent diffusion models.
\newblock In \emph{Proceedings of the IEEE/CVF Conference on Computer Vision and Pattern Recognition (CVPR)}, 2022.

\bibitem[Saharia et~al.(2022)Saharia, Chan, Saxena, Li, Whang, Denton, Ghasemipour, Gontijo~Lopes, Karagol~Ayan, Salimans, et~al.]{saharia2022photorealistic}
Chitwan Saharia, William Chan, Saurabh Saxena, Lala Li, Jay Whang, Emily~L Denton, Kamyar Ghasemipour, Raphael Gontijo~Lopes, Burcu Karagol~Ayan, Tim Salimans, et~al.
\newblock Photorealistic text-to-image diffusion models with deep language understanding.
\newblock \emph{Advances in neural information processing systems}, 35:\penalty0 36479--36494, 2022.

\bibitem[Singer et~al.(2022)Singer, Polyak, Hayes, Yin, An, Zhang, Hu, Yang, Ashual, Gafni, Parikh, Gupta, and Taigman]{make}
Uriel Singer, Adam Polyak, Thomas Hayes, Xi Yin, Jie An, Songyang Zhang, Qiyuan Hu, Harry Yang, Oron Ashual, Oran Gafni, Devi Parikh, Sonal Gupta, and Yaniv Taigman.
\newblock Make-a-video: Text-to-video generation without text-video data.
\newblock 2022.

\bibitem[Sitzmann et~al.(2021)Sitzmann, Rezchikov, Freeman, Tenenbaum, and Durand]{sitzmann2021light}
Vincent Sitzmann, Semon Rezchikov, Bill Freeman, Josh Tenenbaum, and Fredo Durand.
\newblock Light field networks: Neural scene representations with single-evaluation rendering.
\newblock \emph{Advances in Neural Information Processing Systems}, 34:\penalty0 19313--19325, 2021.

\bibitem[studio(2023)]{morph}
Morph studio, 2023.
\newblock \url{https://app.morphstudio.com/}.

\bibitem[Tuo et~al.(2023)Tuo, Xiang, He, Geng, and Xie]{tuo2023anytext}
Yuxiang Tuo, Wangmeng Xiang, Jun-Yan He, Yifeng Geng, and Xuansong Xie.
\newblock Anytext: Multilingual visual text generation and editing.
\newblock \emph{arXiv preprint arXiv:2311.03054}, 2023.

\bibitem[Van Den~Oord et~al.(2017)Van Den~Oord, Vinyals, et~al.]{van2017neural}
Aaron Van Den~Oord, Oriol Vinyals, et~al.
\newblock Neural discrete representation learning.
\newblock \emph{Advances in neural information processing systems}, 30, 2017.

\bibitem[Wang et~al.(2024{\natexlab{a}})Wang, Huang, Shi, Bian, Song, Liu, and Li]{wang2024animatelcm}
Fu-Yun Wang, Zhaoyang Huang, Xiaoyu Shi, Weikang Bian, Guanglu Song, Yu Liu, and Hongsheng Li.
\newblock Animatelcm: Accelerating the animation of personalized diffusion models and adapters with decoupled consistency learning.
\newblock \emph{arXiv preprint arXiv:2402.00769}, 2024{\natexlab{a}}.

\bibitem[Wang et~al.(2023{\natexlab{a}})Wang, Yuan, Chen, Zhang, Wang, and Zhang]{wang2023modelscope}
Jiuniu Wang, Hangjie Yuan, Dayou Chen, Yingya Zhang, Xiang Wang, and Shiwei Zhang.
\newblock Modelscope text-to-video technical report.
\newblock \emph{arXiv preprint arXiv:2308.06571}, 2023{\natexlab{a}}.

\bibitem[Wang et~al.(2024{\natexlab{b}})Wang, Yuan, Zhang, Chen, Wang, Zhang, Shen, Zhao, and Zhou]{wang2024videocomposer}
Xiang Wang, Hangjie Yuan, Shiwei Zhang, Dayou Chen, Jiuniu Wang, Yingya Zhang, Yujun Shen, Deli Zhao, and Jingren Zhou.
\newblock Videocomposer: Compositional video synthesis with motion controllability.
\newblock \emph{Advances in Neural Information Processing Systems}, 36, 2024{\natexlab{b}}.

\bibitem[Wang et~al.(2023{\natexlab{b}})Wang, Chen, Ma, Zhou, Huang, Wang, Yang, He, Yu, Yang, et~al.]{wang2023lavie}
Yaohui Wang, Xinyuan Chen, Xin Ma, Shangchen Zhou, Ziqi Huang, Yi Wang, Ceyuan Yang, Yinan He, Jiashuo Yu, Peiqing Yang, et~al.
\newblock Lavie: High-quality video generation with cascaded latent diffusion models.
\newblock \emph{arXiv preprint arXiv:2309.15103}, 2023{\natexlab{b}}.

\bibitem[Wang et~al.(2023{\natexlab{c}})Wang, Yuan, Wang, Chen, Xia, Luo, and Shan]{wang2023motionctrl}
Zhouxia Wang, Ziyang Yuan, Xintao Wang, Tianshui Chen, Menghan Xia, Ping Luo, and Ying Shan.
\newblock Motionctrl: A unified and flexible motion controller for video generation.
\newblock \emph{arXiv preprint arXiv:2312.03641}, 2023{\natexlab{c}}.

\bibitem[Wu et~al.(2022)Wu, Liang, Ji, Yang, Fang, Jiang, and Duan]{wu2022nuwa}
Chenfei Wu, Jian Liang, Lei Ji, Fan Yang, Yuejian Fang, Daxin Jiang, and Nan Duan.
\newblock N{\"u}wa: Visual synthesis pre-training for neural visual world creation.
\newblock In \emph{European conference on computer vision}, pages 720--736. Springer, 2022.

\bibitem[Wu et~al.(2023{\natexlab{a}})Wu, Ge, Wang, Lei, Gu, Shi, Hsu, Shan, Qie, and Shou]{wu2023tune}
Jay~Zhangjie Wu, Yixiao Ge, Xintao Wang, Stan~Weixian Lei, Yuchao Gu, Yufei Shi, Wynne Hsu, Ying Shan, Xiaohu Qie, and Mike~Zheng Shou.
\newblock Tune-a-video: One-shot tuning of image diffusion models for text-to-video generation.
\newblock In \emph{Proceedings of the IEEE/CVF International Conference on Computer Vision}, pages 7623--7633, 2023{\natexlab{a}}.

\bibitem[Wu et~al.(2023{\natexlab{b}})Wu, Chen, Yang, Guo, Li, and Zhang]{wu2023lamp}
Ruiqi Wu, Liangyu Chen, Tong Yang, Chunle Guo, Chongyi Li, and Xiangyu Zhang.
\newblock Lamp: Learn a motion pattern for few-shot-based video generation.
\newblock \emph{arXiv preprint arXiv:2310.10769}, 2023{\natexlab{b}}.

\bibitem[Yang et~al.(2024)Yang, Gui, Yuan, Liang, Ding, Hu, and Chen]{yang2024glyphcontrol}
Yukang Yang, Dongnan Gui, Yuhui Yuan, Weicong Liang, Haisong Ding, Han Hu, and Kai Chen.
\newblock Glyphcontrol: Glyph conditional control for visual text generation.
\newblock \emph{Advances in Neural Information Processing Systems}, 36, 2024.

\bibitem[Yin et~al.(2023)Yin, Wu, Liang, Shi, Li, Ming, and Duan]{yin2023dragnuwa}
Shengming Yin, Chenfei Wu, Jian Liang, Jie Shi, Houqiang Li, Gong Ming, and Nan Duan.
\newblock Dragnuwa: Fine-grained control in video generation by integrating text, image, and trajectory.
\newblock \emph{arXiv preprint arXiv:2308.08089}, 2023.

\bibitem[Zhang et~al.(2023{\natexlab{a}})Zhang, Wu, Liu, Zhao, Ran, Gu, Gao, and Shou]{zhang2023show}
David~Junhao Zhang, Jay~Zhangjie Wu, Jia-Wei Liu, Rui Zhao, Lingmin Ran, Yuchao Gu, Difei Gao, and Mike~Zheng Shou.
\newblock Show-1: Marrying pixel and latent diffusion models for text-to-video generation.
\newblock \emph{arXiv preprint arXiv:2309.15818}, 2023{\natexlab{a}}.

\bibitem[Zhang et~al.(2023{\natexlab{b}})Zhang, Rao, and Agrawala]{zhang2023adding}
Lvmin Zhang, Anyi Rao, and Maneesh Agrawala.
\newblock Adding conditional control to text-to-image diffusion models.
\newblock In \emph{Proceedings of the IEEE/CVF International Conference on Computer Vision}, pages 3836--3847, 2023{\natexlab{b}}.

\bibitem[Zhang et~al.(2023{\natexlab{c}})Zhang, Wang, Zhang, Zhao, Yuan, Qin, Wang, Zhao, and Zhou]{zhang2023i2vgen}
Shiwei Zhang, Jiayu Wang, Yingya Zhang, Kang Zhao, Hangjie Yuan, Zhiwu Qin, Xiang Wang, Deli Zhao, and Jingren Zhou.
\newblock I2vgen-xl: High-quality image-to-video synthesis via cascaded diffusion models.
\newblock \emph{arXiv preprint arXiv:2311.04145}, 2023{\natexlab{c}}.

\bibitem[Zhao et~al.(2023)Zhao, Gu, Wu, Zhang, Liu, Wu, Keppo, and Shou]{zhao2023motiondirector}
Rui Zhao, Yuchao Gu, Jay~Zhangjie Wu, David~Junhao Zhang, Jiawei Liu, Weijia Wu, Jussi Keppo, and Mike~Zheng Shou.
\newblock Motiondirector: Motion customization of text-to-video diffusion models.
\newblock \emph{arXiv preprint arXiv:2310.08465}, 2023.

\bibitem[Zheng et~al.(2024)Zheng, Peng, and You]{opensora}
Zangwei Zheng, Xiangyu Peng, and Yang You.
\newblock Open-sora: Democratizing efficient video production for all, 2024.

\bibitem[Zhou et~al.(2022)Zhou, Wang, Yan, Lv, Zhu, and Feng]{zhou2022magicvideo}
Daquan Zhou, Weimin Wang, Hanshu Yan, Weiwei Lv, Yizhe Zhu, and Jiashi Feng.
\newblock Magicvideo: Efficient video generation with latent diffusion models.
\newblock \emph{arXiv preprint arXiv:2211.11018}, 2022.

\end{thebibliography}
}


\end{document}